\documentclass[letterpaper, 10 pt, conference]{ieeeconf}
\IEEEoverridecommandlockouts
\usepackage[utf8]{inputenc}
\usepackage{amsmath,amssymb,amsfonts}
\usepackage{blindtext, graphicx}
\PassOptionsToPackage{greek,english}{babel}
\usepackage{textcomp}
\usepackage{xcolor}
\usepackage{booktabs} 
\usepackage{bbding}
\usepackage{soul,color}
\usepackage{comment} 
\usepackage{kantlipsum}
\usepackage{multicol}
\usepackage{microtype}
\usepackage{array}
\usepackage{tabulary}
\usepackage{chngcntr}
\usepackage{multirow}
\usepackage[T1]{fontenc}
\usepackage{amssymb}

\usepackage{tablefootnote}
\usepackage{threeparttable}

\usepackage{cleveref}

\let\Setlength\setlength 
\usepackage{calc}
\newlength{\arrayrulewidthOriginal}

\DeclareMathAlphabet\mathzapf{T1}{pzc}{mb}{it}

\usepackage[style=ieee,minbibnames=1,maxbibnames=2,doi=false,isbn=false,url=false,eprint=false,date=year,bibencoding=utf8]{biblatex}
\AtEveryBibitem{\clearfield{pages}}
\AtEveryBibitem{\clearfield{volume}}
\AtEveryBibitem{\clearfield{number}}
\AtEveryBibitem{\clearfield{note}}
\AtEveryBibitem{\clearlist{language}}

\newcolumntype{K}[1]{>{\centering\arraybackslash}p{#1}}
\usepackage{float}
\usepackage[linesnumbered,ruled,vlined]{algorithm2e}
\usepackage{bm}

\newcolumntype{M}[1]{>{\centering\arraybackslash}m{#1}}
\Setlength{\intextsep}{5pt}
\renewcommand{\baselinestretch}{1} 

\IEEEoverridecommandlockouts                              

\begin{document}
\title{\textbf{Multi-Centroid Hyperdimensional Computing Approach \\ for Epileptic Seizure Detection}
\thanks{This work has been partially supported by the ML-Edge Swiss National Science Foundation (NSF) Research project (GA No. 200020182009/1), and the PEDESITE Swiss NSF Sinergia project (GA No. SCRSII5 193813/1).}
\thanks{$^{1}$ U. Pale, T. Teijeiro, and D. Atienza are with the Embedded Systems Laboratory (ESL) of Swiss Federal Institute of Technology Lausanne (EPFL), Switzerland.\newline
{\tt\footnotesize \{una.pale, tomas.teijeiro, david.atienza\}@epfl.ch}}
}


\author{ Una Pale, Tomas Teijeiro, and David Atienza $^{1}$ }

\maketitle

\begin{abstract}
Long-term monitoring of patients with epilepsy presents a challenging problem from the engineering perspective of real-time detection and wearable devices design. It requires new solutions that allow continuous unobstructed monitoring and reliable detection and prediction of seizures. 
A high variability in the electroencephalogram (EEG) patterns exists among people, brain states, and time instances during seizures, but also during non-seizure periods. This makes epileptic seizure detection very challenging, especially if data is grouped under only seizure and non-seizure labels. 

Hyperdimensional (HD) computing, a novel machine learning approach, comes in as a promising tool. However, it has certain limitations when the data shows a high intra-class variability. Therefore, in this work, we propose a novel semi-supervised learning approach based on a multi-centroid HD computing. The multi-centroid approach allows to have several prototype vectors representing seizure and non-seizure states, which leads to significantly improved performance when compared to a simple 2-class HD model. 

Further, real-life data imbalance poses an additional challenge and the performance reported on balanced subsets of data is likely to be overestimated. Thus, we test our multi-centroid approach with three different dataset balancing scenarios, showing that performance improvement is higher for the less balanced dataset. More specifically, up to 14\% improvement is achieved on an unbalanced test set with 10 times more non-seizure than seizure data. 
At the same time, the total number of sub-classes is not significantly increased compared to the balanced dataset. Thus, the proposed multi-centroid approach can be an important element in achieving a high performance of epilepsy detection with real-life data balance or during online learning, where seizures are infrequent.
\end{abstract}


\section{Introduction}
\bstctlcite{IEEEexample:BSTcontrol}

Epilepsy is a chronic neurological disorder characterized by the unpredictable occurrence of seizures. It is a challenging problem, both from the engineering aspects of real-time detection and wearable devices design, as well as medical aspects. It impacts a significant portion of the world population (0.6 to 0.8\%)~\cite{mormann_seizure_2007}, out of which one-third of patients still suffer from seizures despite pharmacological treatments~\cite{schmidt_evidence-based_2012}. The unexpected occurrence of seizures imposes serious health risks and many restrictions on daily life. Thus, there is a clear need for solutions that allow continuous unobstructed monitoring and reliable detection (and ideally prediction) of seizures. Moreover, these solutions will further be instrumental for designing novel treatments, hence, assisting patients in their daily lives and preventing possible accidents.

In this context, different wearable devices for epilepsy monitoring have been proposed in the literature (e.g., \cite{poh_continuous_2010, beniczky_detection_2013, sopic_e-glass:_2018}). However, there is still a long road ahead to create smart and low-power devices that are well accepted by the medical community and the patients. One of the biggest challenges is to achieve a high enough sensitivity with few or no false positives while considering the vast disbalance in data distribution (i.e., the amount of seizure vs. non-seizure data). Also, another key challenge is related to the usability and comfortability of the wearable device, as it needs to be lightweight, non-stigmatizing, and with extensive battery life. This makes many state-of-the-art algorithms for epilepsy detection 
~\cite{emami_seizure_2019, ghosh-dastidar_principal_2008} infeasible due to excessive memory and/or power requirements.  

Given the aforementioned challenges, Hyperdimensional (HD) computing comes as an interesting alternative. It has lower energy and memory requirements~\cite{burrello_ensemble_2021,asgarinejad_detection_2020}, and there have been hardware implementations and optimizations adapted for it that show promising results~\cite{burrello_ensemble_2021,asgarinejad_detection_2020}.
HD computing is based on computations with very long vectors (usually >10000 dimensions), which represent information in a condensed way. The inspiration for data representation in the shape of long and redundant (mostly binary) vectors came from the neuroscience research. The research stated the  hypothesis that the brain's computation is based on the high-dimensional randomized representation of data rather than scalar numerical values~\cite{kanerva_hyperdimensional_2009}.

\begin{figure*}[]
    \centering
    \vspace{2mm}
    \includegraphics[width=\linewidth]{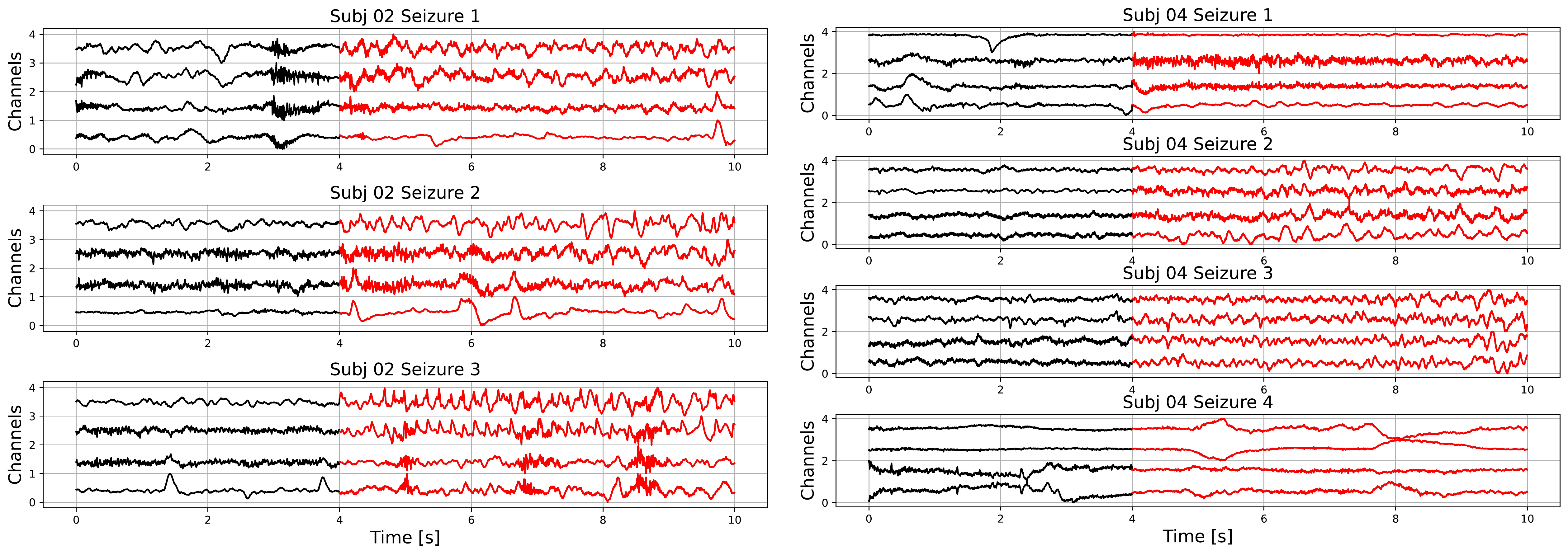}
    \caption{\small{Raw signal showing several seizures from subjects 2 and 4 of the CHB-MIT database. Only the first 4 channels are shown.)}} 
    \label{fig:rasData}
    \vspace{-4mm}
\end{figure*}

HD computing is an entirely different approach to machine learning (ML) than most other state-of-the-art algorithms. It is based on mapping data and its relations in the form of long vectors, followed by the relatively simple process of learning and inferring predictions from them. 
In particular, its three main stages are encoding, training, and querying. First, baseline vectors representing different scalar values are combined during the encoding stage to represent data (either raw data or features). This process leads to one single vector representing each data sample instead of, as in other ML approaches, a feature set representing it. Next, during training, all vectors from the same class are summed up (bundled) to one prototype vector representing each class. In the end, for inference, prototype vectors of all classes are compared with the current data instance vector, and the label of the most similar one is given as output. 

Such an encoding, learning, and inferring approach enables many new possibilities compared to standard ML. In fact, its low computational and memory requirements ~\cite{burrello_ensemble_2021,asgarinejad_detection_2020} make it interesting for low-power small-size wearable devices. For example, in recent years, a lot of effort has been put into designing wearable devices for patient monitoring, with detection and prediction capabilities. One of such applications is epilepsy monitoring and real-time seizures detection. 

HD computing is exciting due to the several opportunities it offers. One of them is, for example, continuous learning ~\cite{imani_semihd_2019}, which is easily implementable due to the simplicity of the training procedures of HD computing. This is relevant for epileptic seizure detection due to the inherent scarcity of epileptic seizure recordings, thus, the small amount of seizure training data available. An additional opportunity is the use of semi-supervised learning approaches with HD computing. In the literature, the form of iterative learning~\cite{imani_adapthd_2019}) has been proposed, but it has not yet been fully explored for epilepsy. However, unsupervised or at least semi-supervised learning would be very useful due to the time-consuming and complex process of labeling data. Further, HD computing can enable a closer interaction between personalized and generalized models, being an option for distributed learning~\cite{imani_framework_2019}. 

Traditionally, HD computing classifiers have been based on creating one model vector for each target class. However, a challenging aspect of electroencephalogram (EEG) signatures of epileptic seizures is their uniqueness and high variability among people, brain states, and time instances, especially if they are grouped under only two given labels (seizure and non-seizure). Further, non-seizure data also contains many different brain states, such as awake, sleeping, physical or mental effort conditions, etc. All of these states have their own brain signatures. Thus, we hypothesize in this work that creating multiple sub-types (model vector centroids) of seizure and non-seizure classes, based on both labels provided by a neurologist and also on EEG signal characteristics, can be more appropriate.  

Following the previous observations about epilepsy, we present a novel semi-supervised learning approach for HD computing to evaluate whether a multi-centroid representation of the seizure and non-seizure states can improve seizure detection performance. 
More precisely, in this work, we contribute to state of the art in the following manner:
\begin{itemize}
    \item We design a semi-supervised HD computing approach of learning based on the unconstrained creation of several prototype vectors/sub-classes (unlabeled) of main (labeled) classes.
    \item We implement this novel approach for epileptic seizure detection based on EEG signal recordings, leading to the creation of multiple prototype vectors/sub-classes for seizure and non-seizure. We evaluate and show a significant improvement in the performance when compared to the standard 2-class HD approach.
    \item Since a high number of prototype vectors penalizes memory efficiency, we designed two algorithms to reduce the number of sub-classes in the post-training stage. One is based on removing less populated sub-classes, and the other is based on clustering of sub-classes.  
    \item We measure the performance improvement of this approach and analyze the number and structure of sub-classes based on the publicly available CHB-MIT epilepsy database. We show that this approach has greater improvements for more unbalanced datasets while not significantly increasing the number of sub-classes compared to balanced datasets. Thus, this multi-centroid approach can be an essential element to achieve high performance of epilepsy detection with real-life data structures. Moreover, it can be particularly relevant during online learning, where seizures are infrequent. 
\end{itemize}

\begin{figure*}[]
    \centering
    \vspace{2mm}
    \includegraphics[width=\linewidth]{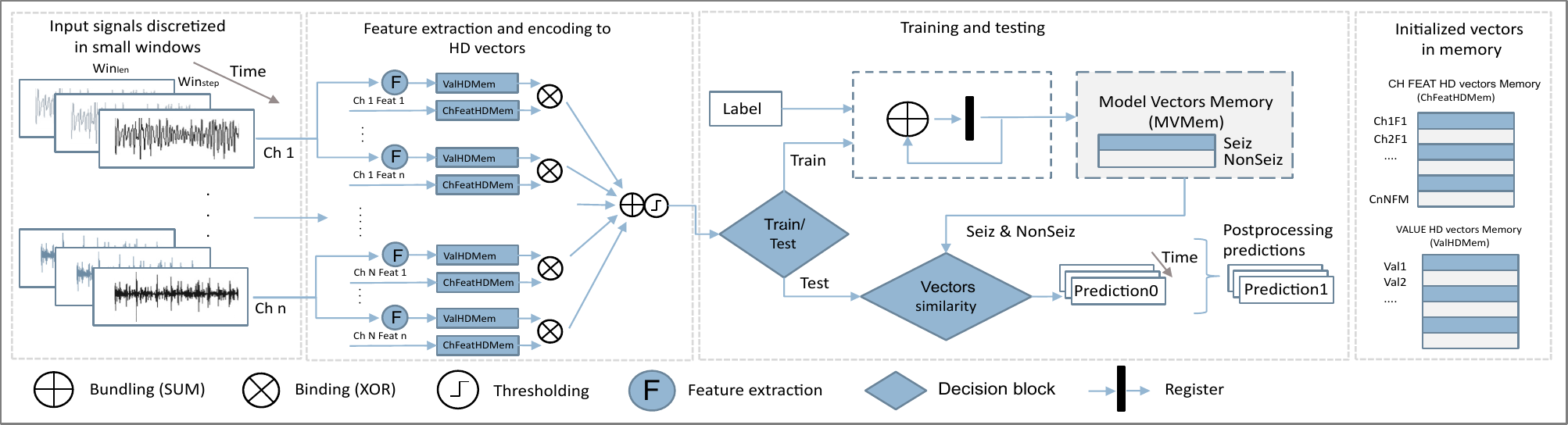}
    \caption{\small{Schematic of the HD computation workflow}} 
    \label{fig:workflowSchematic}
    \vspace{-4mm}
\end{figure*}

\section{Background and Related Work}

HD computing is based on few specific algebraic properties when computing with HD vectors. First, any randomly chosen pair of vectors is nearly orthogonal. Second, if we sum two or more vectors, the result will be with high probability more similar to the added vectors than to any other randomly chosen vector. 
The most common subtype of used vectors are binary ones, where their elements can be only 0 or 1. In practice, also tertiary (-1,0,1) or integer/float vectors are sometimes used. Summation of the vectors is usually done by bit-wise summation with majority voting normalization. 

Representing data as HD vectors enables simple training procedures for classification problems, where all vectors from the same class can be summed up (and normalized) to represent a prototype vector of that class. Later, during the prediction process, the similarity between an HD vector representing the current sample and the prototype HD vectors for each class is calculated, and the label of the most similar prototype vector is given. The similarity is measured as the distance between two vectors, which can be the Hamming distance for binary vectors or cosine (or dot) product for integer, or floating-point value vectors.

HD computing has been applied for different challenges in the domain of biomedical applications: EEG error related potentials detection~\cite{rahimi_hyperdimensional_2020}, electromyogram (EMG), gesture recognition~\cite{rahimi_hyperdimensional_2016}, emotion recognition from GSR (galvanic-skin response), electrocardiogram (ECG) and EEG~\cite{chang_hyperdimensional_2019}, etc. In the specific application of epileptic seizure detection, there are few recent papers that have claimed promising results when using EEG or intracranial EEG (iEEG) data. 

The first paper that applied HD computing to epileptic seizure detection was based on transforming data to local binary patterns (LBPs), which were then mapped to HD vectors~\cite{burrello_one-shot_2018}. LBPs are short binary arrays that represent whether a signal is increasing or decreasing. In~\cite{burrello_one-shot_2018}, authors used iEEG data from patients from the Inselspital Bern epilepsy surgery program and focused on testing one-shot learning, or learning from as few seizure instances as possible. 
Later, the authors extended this work in~\cite{burrello_ensemble_2021} by using also the mean amplitude and line length features, besides LBP, to describe data. Each feature forms its own prototype vector for every class and acts as a standalone classifier. Then, the predictions (more precisely, vector distances) are fed into a single-layer perceptron with three neurons to decide the final prediction. The authors show better performance and lower latency on the same dataset than the previous paper~\cite{burrello_one-shot_2018}. These works also compared and showed advantages over other state-of-the-art algorithms for epilepsy detection regarding performance, memory, and computational requirements. 

In~\cite{asgarinejad_detection_2020}, authors used EEG data, which is more viable for continuous long-term monitoring, and compared HD computing with different standard state-of-the-art ML approaches (KNN, SVM, regression, random forests, and CNN). They used 54 different features from~\cite{zanetti_robust_2020} for KNN, SVM, regression, and random forests, and raw amplitude values of signals encoded to HD vectors for the proposed HD approach. The CHB-MIT database from the Children's Hospital of Boston and MIT~\cite{shoeb_application_2009, goldberger_ary_l_physiobank_2000} was used. The authors reported that their HD approach surpassed the performance of all other approaches.

Since in many recent HD papers, various approaches to map data (or features) to HD vectors were used; in~\cite{pale_systematic_2021} the authors compared several approaches for the task of epileptic seizure detection. They present in detail different methods of mapping data to HD vectors, namely, LBP/raw data, frequency composition (FFT), single feature, or any number of features. They show significant differences in performance as well as memory and computation requirements between them.  

Even though current papers applying HD computing for epilepsy show very promising results, they are still quite far from real-life applications due to various data preparation and selection limitations. 
In particular, most of the papers use only a small portion of the data available in the databases for training, most often balancing the amount of seizure and non-seizure data. However, this context is very far from the actual seizure - non-seizure ratio in a real-life scenario. 

In addition, results are very sensitive to which data is used for training and testing. This is due to the high variability of seizures even within one subject. For example, in Fig.~\ref{fig:rasData} we show different seizure segments from the same person (subjects 2 and 4 from the CHB-MIT database). A seizure signal can show a very different morphology between different channels of the same seizure, as well as between different seizure instances of the same person. 
Furthermore, non-seizure data can represent many different types of neural activity, such as resting, mental activity, sleeping, etc. Therefore, it is not realistic to expect to represent it with only one prototype vector.

In this work, due to the intrinsic variability of seizures and non-seizure background data, we hypothesize that creating more prototype vectors for seizures and non-seizures during training can be beneficial for learning and prediction. We called this approach "multi-centroid", as we allow to have more vectors (centroids) representing sub-types of each class. This is a form of semi-supervised learning, as the main labels (seizure or non-seizure) are known, but an unrestricted number of seizure (and non-seizure) sub-types is created during training. 

In the literature, few papers are applying different semi-supervised learning and clustering approaches to HD computing. In~\cite{imani_semihd_2019}, authors allow iterative expanding of the training data by labeling unlabeled data points, which can be classified with high confidence by the current model. This improves the quality of prediction by 10.2\% on average on 18 popular datasets. This approach can be highly beneficial for epilepsy due to the high amount of unlabeled data that can be accumulated during patient monitoring but cannot be fully labeled by the experts. A potential problem is that it would strengthen common patterns, but could under-represent less common patterns. 

In~\cite{imani_hdcluster_2019}, the k-means algorithm is adapted to the HD computing paradigm. This means that, before clustering, data is mapped to HD vectors. Then, properties of HD vectors are used to perform clustering on a preset number of classes. The authors compared it with k-means on 9 different datasets, and the influence of various parameters was investigated. Results showed the same or better performance than the standard non-HD k-means algorithm for all datasets. The disadvantage of this approach is that the number of sub-classes has to be preset, which can be quite challenging in the case of an epileptic seizures. In fact, this number would be different for every patient and it may even change in time as more training data is added. Further, it does not use the information about the global (seizure/non-seizure) labels that are available.  

Another approach for semi-supervised learning is the idea of relearning, in which the algorithm iteratively passes through the training set. In the case of a mis-classification, the sample is removed from the mis-classified class and added again to the prototype vector of the correct class. Therefore, iterative learning tries to overcome the problem of single-pass learning that it can lead to the saturation of the prototype vectors of each class by data that are more common in each class and perform badly on under-represented patterns of the same class. In~\cite{imani_adapthd_2019}, authors tested iterative approaches with different fixed and adaptive learning rates on several datasets for speeding up learning and saving energy while keeping the same or higher accuracy as single-pass training. 

In~\cite{hernandez-cano_real-time_2021}, the authors targeted to achieve the higher performance of iterative training while keeping the speed and simplicity of single-pass training. The approach, called OnlineHD, is single-pass, but adjusts the weight of each example according to the similarity with the trained prototype vectors. 
This leads to an accuracy increase of 12.1\% in average, when compared to single-pass HD approaches, and has 13x fewer iterations on average than iterative HD approaches. 

In the scope of this paper, conversely to previous works, we propose a new approach called "multi-centroid". More precisely, if the current data vector is more similar to an incorrect class than to any of the correct sub-classes, we create a new sub-class of the correct class. In this way, less common data patterns will have their own sub-class and will not get over-voted and under-represented by more common patterns. This semi-supervised approach is guided by labeled data, but allows the creation of an unlimited number of sub-classes for each of the main classes. The number of sub-classes is highly dependent on the subject, data training instances, and also the amount of training data, and as such it would be hard to predict and set at the beginning as in~\cite{imani_hdcluster_2019}. 

Our proposed approach has a similar underlying idea as OnlineHD~\cite{hernandez-cano_real-time_2021}, or iterative learning~\cite{imani_adapthd_2019} in that it focuses on less common patterns. However, it is different since it allows the creation of sub-classes rather than adding them multiple times to a single vector. Consequently, our approach enables more control over the classification and potential interpretability of the predictions. 

\begin{figure}[]
    \centering
    \vspace{2mm}
\includegraphics[width=\linewidth]{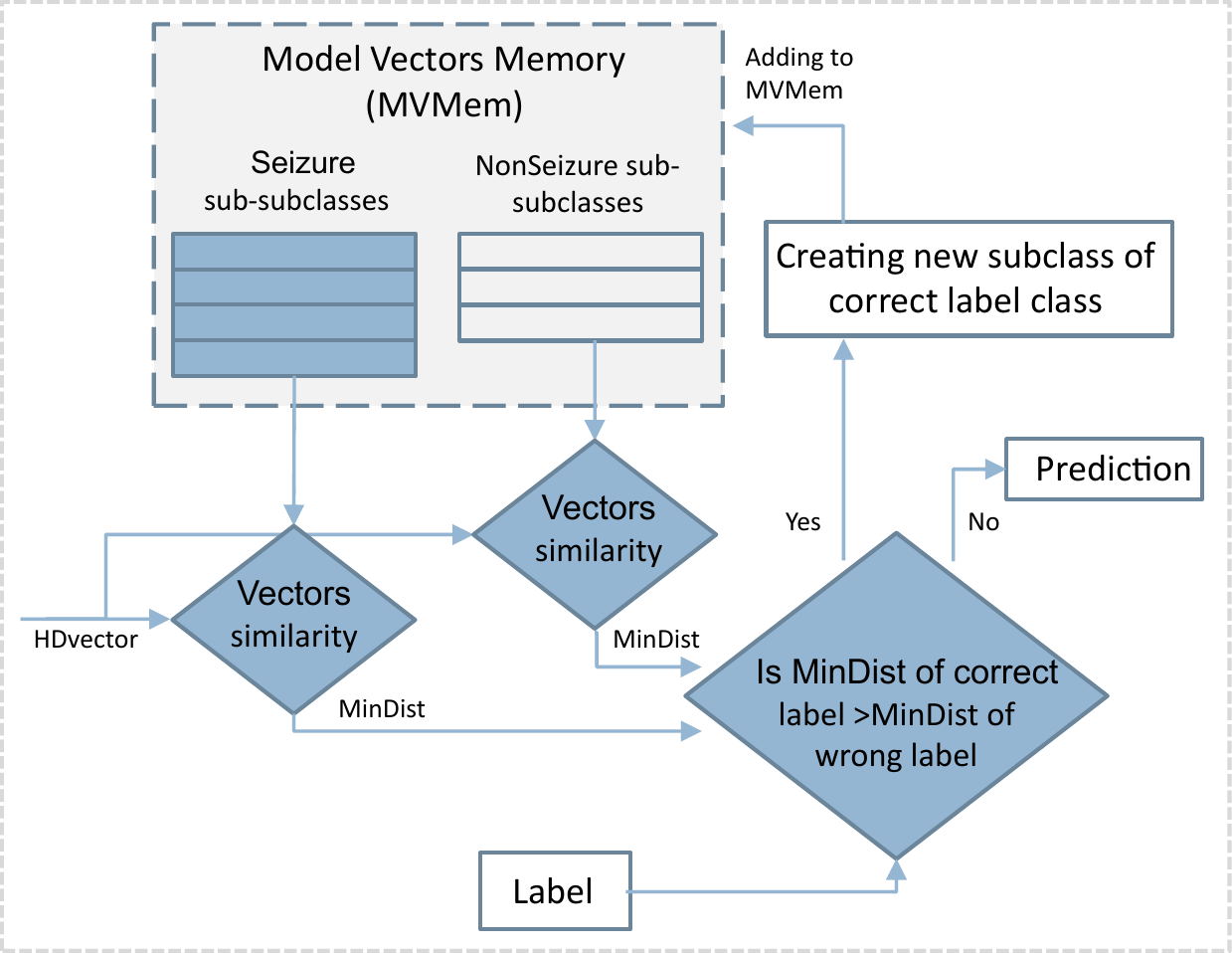}
    \caption{\small{Schematic of the first step of multi-centroid training workflow where new sub-classes are created}} 
    \label{fig:MultiClassWorkflowSchematic}
\end{figure}

\section{Multi-centroid based HD framework and workflow}
\label{sec:OverallWorkflow}

\subsection{Classical HD training and testing workflow} 
\label{Subsec:TrainTestWorkflow}

The classical HD computing analysis workflow is presented in Fig.~\ref{fig:workflowSchematic}. Data is discretized into windows of duration $W_{len}$, for which features are calculated and encoded into an HD vector representing that data instance. This is repeated every $W_{step}$, i.e., a prediction is given based on $W_{len}$ of data every $W_{step}$. In our case, $W_{len}$ has been 8s, and $W_{step}$ 1s.

Before training, vector memory maps need to be initialized. This means that we assign a static vector to each possible feature value, and to each combination of feature and channel. Features are normalized and discretized in the same number of levels, which allows us to use the same vectors to represent the values of all features. Thus, both feature value vectors $HDV_{Val}$, and feature-channel index vectors $HDV_{ChFeat}$  representing a feature of a specific channel, are generated once before the training starts. More precisely, if we have M features and N channels, NxM $HDV_{ChFeat}$ vectors will be initialized. 

$HDV_{ChFeat}$ vectors representing features and channels are independently and randomly generated as there is no specific relation between features and channels. 
On the other hand, $HDV_{Val}$ vectors are initialized in a way where first the vector is randomly initialized. Still, every subsequent vector representing the next possible value is created from the previous one by permuting consecutive blocks of $d$ bits. The number of bits $d$ depends on the number of possible needed values (and corresponding $HDV_{Val}$ vectors). This approach ensures that vectors representing numbers that have closer values are also more similar. 

\begin{figure*}[]
    \centering
    \vspace{2mm}
\includegraphics[width=\linewidth]{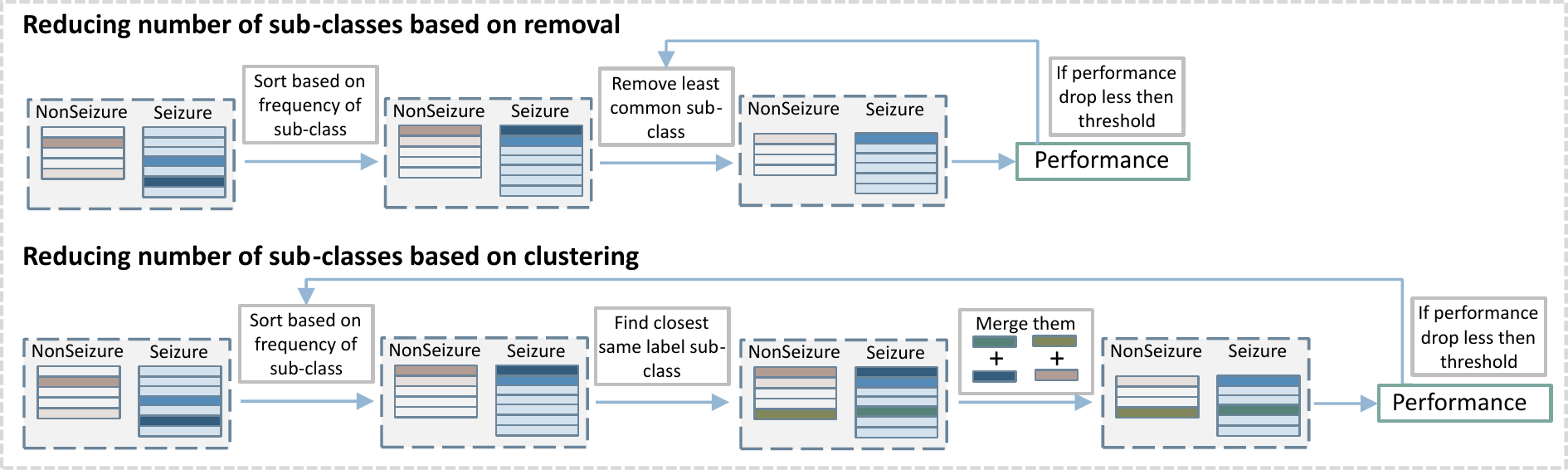}
    \caption{\small{Diagram of the second step of our multi-centroid training workflow where number of sub-classes is reduced}} 
    \label{fig:multiClassreductionPart}
    \vspace{-4mm}
\end{figure*}

During the training step, a feature value vector $HDV_{Val}$ of a specific channel is bound to a pre-initialized vector $HDV_{ChFeat}$ representing the feature of that particular channel. All those bound vectors are then bundled (by summing and normalizing) to get a vector representing the current data window. Next, windows belonging to one class are bundled into one HD prototype vector for that class. Since in our case vectors are binary, binding is performed by means of a XOR operation, while bundling is achieved by performing bit-wise summation (SUM) over the HD vectors and rounding based on majority voting. 

For epileptic seizure detection, specifically, our approach leads in the end to two prototype HD vectors: one for the ictal and one for the interictal class. Ictal relates to the part of the data where the seizure was present. Conversely, interictal corresponds to the baseline EEG data distant from seizure episodes. Around ictal data, often pre-ictal and post-ictal phases are defined, but here we focus only on clear seizure (ictal) and non-seizure (interictal) classification. 

In the testing phase, the HD vector of a specific window is compared with the prototype vectors, and the label of the more similar class is chosen. In our case, we used the Hamming distance to quantify this similarity.

\subsection{Multi-centroid training workflow}
\label{Subsec:MuliClassWorkflow}
\subsubsection{Creation of sub-classes}
\label{Subsec:MuliClassWorkflow1}
The classical single-pass training procedure, as explained in Sec.~\ref{Subsec:TrainTestWorkflow}, has the drawback that all the training samples are equally important and summed up to the same class prototype vector. This leads to a dominance by the most common patterns of the prototype vectors, while less common patterns are under-represented. Thus, in our proposed multi-centroid approach, as illustrated in Fig.~\ref{fig:MultiClassWorkflowSchematic}, we detect the difference of the current pattern/vector from the existing prototype vector. Then, in case of a significant difference, we create a new sub-class with its associated prototype vector. This significant difference is estimated by comparing the current vector with the prototype vectors of all the sub-classes of the correct class and of the wrong classes. If the most similar prototype is from a wrong class, then a new sub-class is created for the correct class, initialized with the current vector. As a result, our approach is also single-pass, thus the training procedure has the same complexity order as the classical one.

\subsubsection{Reducing the number of sub-classes}
\label{Subsec:MuliClassWorkflow}
The first part of our approach creates new sub-classes without any additional constraint, except that new sub-classes have to be significantly different from the existing ones. As Fig.~\ref{fig:PercDataPerSubclasses} shows, this procedure can sometimes result in a number of sub-classes that were created by just a few samples, and which might not contribute to the final prediction performance significantly. At the same time, they increase the memory requirements of the system as memory is linearly related to the number of HD prototype vectors needed to store. Thus, the second part of the algorithm removes some of the sub-classes while still keeping the increased performance benefits. Two methods were tested, as illustrated in Fig.~\ref{fig:multiClassreductionPart}: the first one is based on removing less common sub-classes, and the second is based on clustering them. 

The approach to remove less common sub-classes starts by sorting classes based on the amount of data used to create them during training. Then, it removes them in steps, starting from least populated, while monitoring the performance after each step of removal. The performance is evaluated on the training set, and the iterative removal process is stopped once the performance drops for more than a pre-selected threshold.

Instead of removing less common sub-classes, the second approach merges them with the closest same label sub-class. The process is also performed iteratively. It starts from less common sub-classes, while monitoring the prediction performance and stopping the process in case of a performance drop bigger than a preset threshold.

\section{Experimental Setup}
\label{Sec: Experimental_setup}

\subsection{Databases}
\label{Subsec:databases}

As mentioned previously, our proposed multi-centroid approach is compared with the standard 2-class HD approach using EEG data on the use-case of epileptic seizures detection.  
We use the publicly available CHB-MIT database~\cite{shoeb_application_2009, goldberger_ary_l_physiobank_2000} to prepare three different datasets. Namely, often HD algorithms are tested on balanced versions of the databases where a sample of non-seizure data is randomly selected from raw data and matched in duration with seizure data. This often simplifies computation and performance assessment while also allowing us to focus on the separability of the classes and preventing problems related to the class data distribution. Unfortunately, this does not represent real-life data distribution and can lead to a highly overestimated performance, which cannot be achieved during continuous monitoring with a wearable device. Thus, in this work, we use three different distributions of data: 1) balanced with an equal amount of ictal and interictal data, and 2) and 3) unbalanced with 5 and 10 times more interictal data. We call these three distributions F1, F5, and F10, respectively.

The CHB-MIT database was collected by the Children's Hospital of Boston and MIT from 24 subjects with medically resistant seizures. It is an EEG dataset with a variable amount of channels. To standardize the experiment, we use the 18 channels from an international 10-20 montage that are common to all patients. Overall, the dataset contains in total 183 seizures, with an average of 7.6 $\pm$ 5.8 seizures per subject. 

During the balancing step of data preparation, when randomly selecting the interictal segments of data, we take care of not including data within 1 minute of seizure onset and up to 15 minutes after a seizure, as this data might contain ictal patterns.

\subsection{Feature extraction and mapping to HD vectors}
\label{Subsec:FeaturesUsed}

In standard ML approaches, more features usually lead to performance improvements~\cite{zanetti_robust_2020}, and this has also been shown for HD~\cite{pale_systematic_2021}. Thus, we use the same approach using 45 features as in~\cite{pale_systematic_2021} but with an additional feature of mean amplitude value. The initial 45 features, based on~\cite{zanetti_robust_2020}, contain 37 different entropy features, including sample, permutation, Renyi, Shannon, and Tsallis entropies, as well as 8 features from the frequency domain. 
For frequency-domain features, we compute the power spectral density and extract the relative power in the five common brain wave frequency bands; delta: [0.5-4] Hz, theta: [4-8] Hz, alpha: [8-12] Hz, beta: [12-30] Hz, gamma: [30-45] Hz, and a low-frequency component ([0-0.5] Hz), for each signal window. These features are commonly held to be medically relevant for detecting seizures~\cite{teplan_fundamental_2002}.

Next, for each feature, its value $HDV_{Val}$ and its index vector $HDV_{ChFeat}$ are bound (XOR), to get $HDV_{ValFeat}$ vectors. Finally, to get a final HD vector representing each $W_{len}$, we bundle (sum and round) $HDV_{ValFeat}$ vectors of all features and channels, as shown in Fig.~\ref{fig:workflowSchematic}. In this approach, we do not distinguish between channels and treat them all equally important. 

\subsection{Validation}
\subsubsection{Validation strategy}
\label{Subsec:evaluation}
Due to the subject-specific nature of epileptic seizures and their dynamics, the performance is evaluated on a personalized level. Data for each subject was pre-processed and divided into files, where each file contains one seizure, but the specific amount of non-seizure samples depends on the balancing type (1x, 5x, or 10x). This setting supports a leave-one-seizure-out approach, where the HD model is trained on all but one seizure/file. For example, for a subject with $N_{seiz}$ files (each containing one seizure), we perform $N_{seiz}$ leave-one-out training/test cycles 
and measure the final performance for that subject as the average of all cross-validation iterations. 

Besides measuring the performance of seizure predictions, in this experimental analysis we also consider the number of sub-classes created and kept after the optimization steps of sub-classes removal or clustering, as well as the  the amount of data in them. 

\subsubsection{Performance Evaluation}
\label{Subsec:PerfEvaluation}
The system's performance is quantified using several different measures to capture as much information as possible about predictions. Similarly as proposed in~\cite{ziyabari_objective_2019, shah_validation_2020} and later used in~\cite{pale_systematic_2021}, we measure performance on two levels: 1) episode level, and 2) seizure duration level. Seizure duration is based on standard performance measures, where every sample is equally important and treated independently. The episode metric focuses, on the other side, on correctly detecting seizure episodes, but is less concerned about duration and correct prediction of each sample within seizure. 

For both levels, we measure sensitivity (true positive rate or $TPR$, calculated as $TP/(TP+FN)$), precision (positive predictive value or $PPV$, calculated as $TP/(TP+FP)$) and F1 score ($2*TPR*PPV/(TPR+PPV)$). Metrics on these two levels give us a better insight into the operation of the proposed algorithms. Furthermore, the performance measure often depends on the intended application and plays a big role in the acceptance of the proposed technology. 
Finally, in order to have a single measure for easier comparison of methods, we calculate the geometric mean value of F1 score for episodes ($F1E$) and duration($F1D$) as $F1DEgmean=sqrt(F1D*F1E)$.

\subsubsection{Label post-processing}
\label{Subsec:LabelPostprocessing}
We report the performance measures described in Sec.~\ref{Subsec:PerfEvaluation} for raw predictions. However, assuming that the decision of the classifier can change every second is not realistic from a real-time monitoring perspective. Thus, it is advisable to post-process labels before reporting performance figures. This is due to the viable time properties of the seizures, and also due to the small $W_{len}$ and even smaller $W_{step}$, so that granularity is much smaller than the dynamics of the seizures. For example, it is not reasonable for seizure episodes to last only a few data samples, or if two seizures are very close, they probably belong to the same seizure and are so labeled by neurologists. Thus, as data and predictions are time sequences, we exploit time information to smooth the predictions by going through the predicted labels with a moving average window of a certain size $SW_{len}$ (5s) and performing majority voting.

Finally, we have released all the code and data required to reproduce the presented results as open-source\footnote{https://c4science.ch/source/MultiCentroidHD\_public/}.

\section{Experimental Results}
 
\subsection{Prediction performance }
\label{Subsec:predictionPerformance2}

\begin{figure}[]
    \vspace{2mm}
    \centering
    \includegraphics[width=\linewidth]{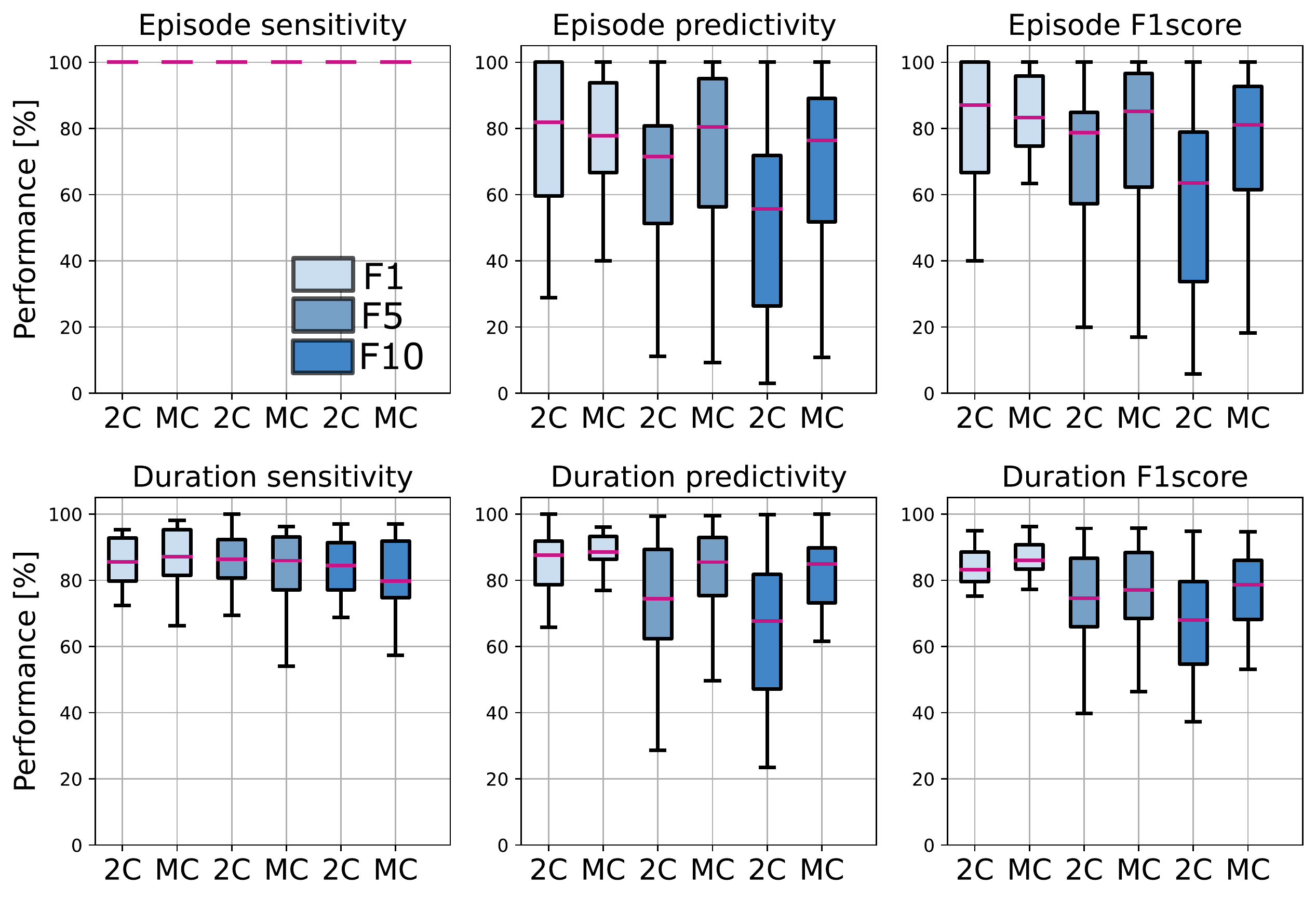}
    \caption{\small{Average performance of all subjects in the test set, for 2 class (2C) and multi-centroid (MC) model. 
    Performance measures shown: sensitivity, predictivity, and $F_1$ score both for episodes, and duration level. 
    }} 
    \label{fig:2CvsMC_6Perf_allFact}
\end{figure}

In Fig.~\ref{fig:2CvsMC_6Perf_allFact}, the performance between 2-class (2C) and multi-centroid (MC) models for all three data balancing cases (F1, F5, F10) is shown. Performance is reported as sensitivity, predictivity, and $F_1$ score for both episode detection and seizure duration detection to get a deeper insight into the performance. 

\begin{figure}[]
    \vspace{2mm}
    \centering
    \includegraphics[width=\linewidth]{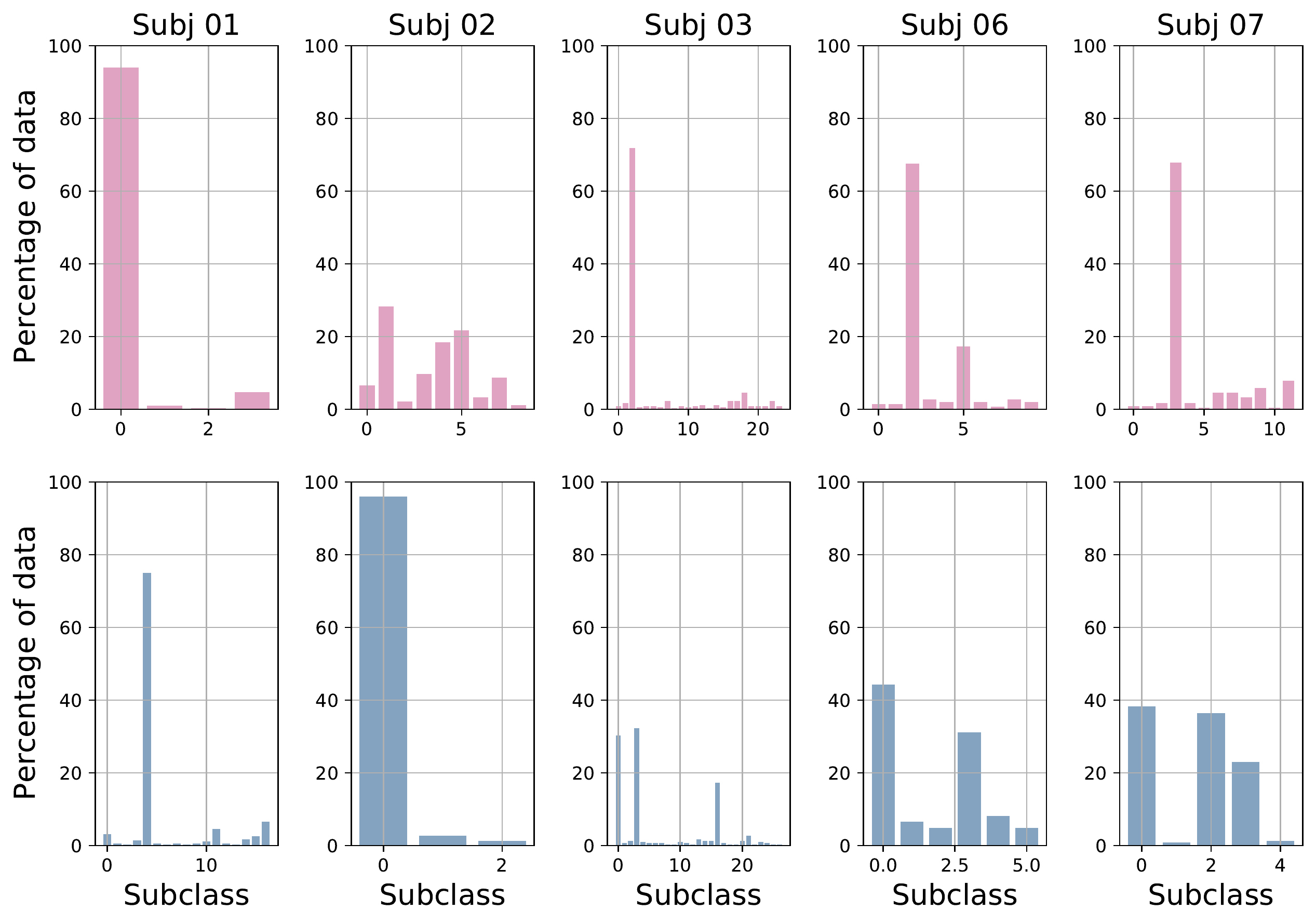}
    \caption{\small{Percentage of data added to each of sub-class, shown for both seizure (red) and non-seizure (blue) sub-classes for 5 randomly selected  subjects.}} 
    \label{fig:PercDataPerSubclasses}
\end{figure}

It is evident that the detection of episodes in the aspect of sensitivity is extremely high even for a 2-class model, so there is no real space for improvement with the multi-class approach. However, predictivity of both episodes and duration of seizures increases with multi-centroid model, 
meaning that less false positives are detected with MC approach than 2C approach. Only for duration sensitivity, even though there is an increase for the training set (not shown here), on the test set, we notice a slight decrease. Therefore, not the whole seizure duration is correctly predicted.
Further, it can be noticed that performance is, in general, worse for non-balanced datasets and that performance drops with more non-seizure data. Therefore, it is required to report all three performance values, as reporting only performance on the balanced dataset (as most works in the literature do) can lead to misleading results about the performance on real-life data distribution. Moreover, the  performance increment due to the multi-centroid approach is higher for more unbalanced datasets (F10 and F5 when compared to F1), which can be explained by the initially higher space for improvement. 

\subsection{Analysis of created sub-classes}
\label{Subsec:subclassesAnalysis}

Fig.~\ref{fig:PercDataPerSubclasses} shows the number and distribution of sub-classes created during multi-centroid training, for both seizure and non-seizure sub-classes for  few subjects. 
First, we can see that the number of sub-classes created is very variable among subjects, some having only a few (e.g., Subj 1 with 4 sub-classes for seizure) and some having a lot of them (e.g., Subj 3 with more than 20 seizure sub-classes). The number of seizure and non-seizure sub-classes is also very variable within the subject. For example, subject 1 has 17 non-seizure sub-classes and only four seizure sub-classes, while subject 7 has 11 seizure sub-classes and only five non-seizure ones. This situation reflects the variability of raw data and demonstrates the rationale for our multi-centroid approach instead of grouping all seizures into one vector (class) and all non-seizures to another vector, as done in the 2-class model. 

Furthermore, it is very interesting to observe the amount of data used to create each of the sub-classes. This corresponds to the frequency of occurrence of each sub-class and shows that there are usually 1 to 3 sub-classes that are very common, while the rest are less common. This also varies greatly between patients, and whether it is a seizure or non-seizure class. 
This is the main motivation behind the strategies we implemented to reduce the number of sub-classes.

\subsection{Reduction of sub-classes}
\label{Subsec:subclassReduction}

\begin{figure}[]
    \vspace{2mm}
    \centering
    \includegraphics[width=\linewidth]{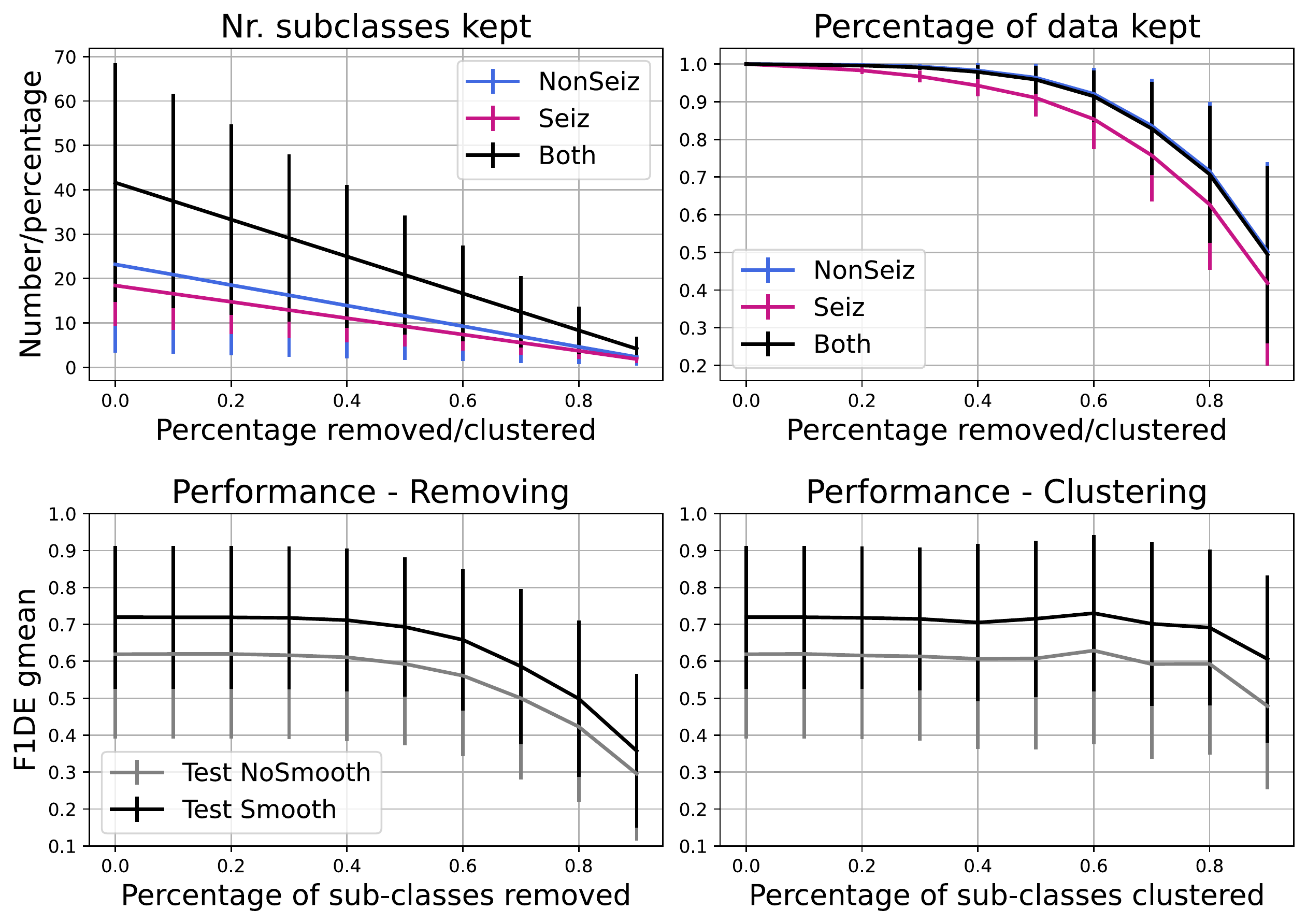}
    \caption{\small{Iterative reduction of sub-classes and how it affects the number of sub-classes and percentage of data in them. Further, performance decrease through steps due to removing/clustering of sub-classes is shown. Performance for test set before (gray) and after (black) smoothing is shown. }} 
    \label{fig:ReducSubclass_graph1}
\end{figure}

In Fig.~\ref{fig:ReducSubclass_graph1}, we show the results of the experiment, where we iteratively remove 10\% of less common sub-classes in every step of the iteration. We see that the number of seizure and non-seizure sub-classes is linearly dropping, while the percentage of data retained is slowly dropping at the beginning and faster later. More specifically, it is reduced more quickly for seizure data as sub-classes are more evenly populated than non-seizure sub-classes. In total, significant data reduction begins to occur once >50\% of sub-classes are removed. 

For the same experiment, w show how the performance (F1DEgmean) decreases while iteratively removing or clustering sub-classes.  
Similarly, there is no significant drop in performance up to 50\% of sub-classes being removed, and after it drops very steeply. The decrease is less steep for clustering and allows even 80\% of sub-classes to be clustered while keeping high performance (meaning <5\% of gmean of $F_1$ score for episodes and duration performance drop).

\subsection{Optimizing performance and number of sub-classes}
\label{Subsec:optPerfandNumSubclasses}

\begin{figure}[]
    \vspace{2mm}
    \centering
\includegraphics[width=\linewidth]{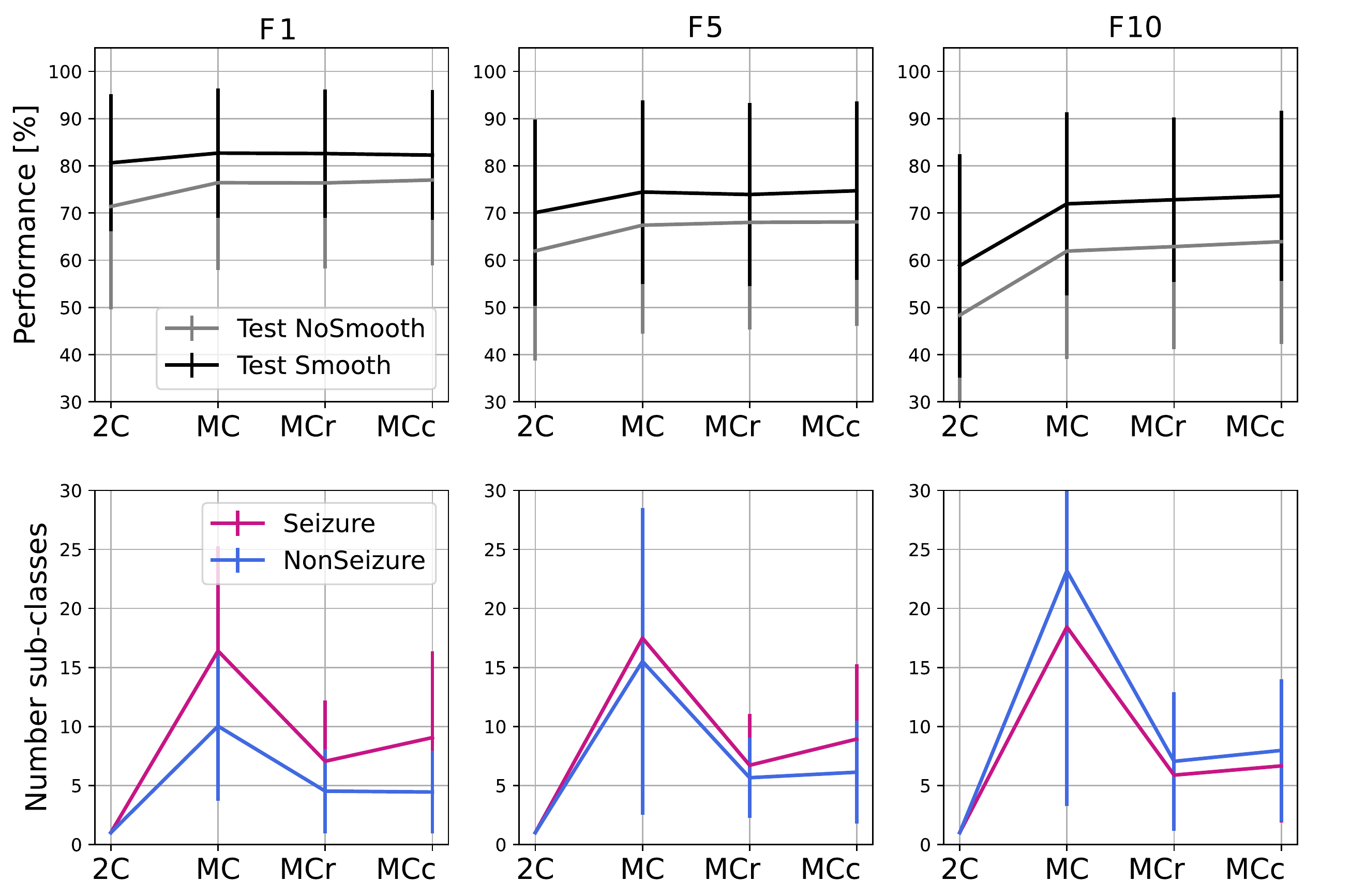}
    \caption{\small{Performance and number of sub-classes for 2-class (2C),  multi-class (MC) approach and with 2 methods for reduction of number of sub-classes: sub-classes removal (MCr) and clustering (MCc).}} 
    \label{fig:ResultsSummary_graph1}
\end{figure}

As shown in Fig.~\ref{fig:ReducSubclass_graph1}, it is possible to reduce the number of sub-classes significantly, while not sacrificing much in terms of performance. Thus, as explained in Sec.~\ref{Subsec:MuliClassWorkflow}, we tested two approaches. The first approach (MCr) removes the less common sub-classes iteratively in steps (10\% of sub-classes in each step) and, after each iteration, evaluates performance on both training and test set. If the performance on the training set drops more than a given tolerance threshold (in this case, 3\% of F1DEgmean was used), the process is stopped and the number of sub-classes is considered optimal.
The second approach (MCc) is based on clustering the less common sub-classes rather than completely removing them. In this approach, as well in iterative steps, we pick the less common sub-classes (10\% of them in each step) and merge them with the most similar sub-classes of the same global label to each of them. The process stops after performance drops more than a tolerance threshold in the training set, the same as in the MCred approach.  
In Fig.~\ref{fig:ResultsSummary_graph1} we show performances and number of sub-classes for 2-class model (2C), initial multi-centroid (MC) model, and after two approaches for optimization, with sub-classes reduction (MCr) and clustering (MCc). Only one performance is shown, F1DEgmean, to simplify comparisons. Finally, we report the results for the test set using all three balancing scenarios (F1, F5, F10).

Based on the results in Fig.~\ref{fig:ResultsSummary_graph1}, we can conclude that the approaches to optimize the number of sub-classes do not significantly degrade performance when compared to the multi-centroid model, which is still substantially higher than the 2-class model (for F5 and F10). P values of Wilcoxon statistical test for difference between 2C and MC approach (after smoothing)  were 0.159, 0.009 and 1.19e-16 for F1, F5 and F10 data balancing, respectively. On the other hand, the number of sub-classes is much smaller when compared to the multi-centroid model. Even though the iterative process and number of sub-classes were decided based on training data performance, test data performance also remains equivalent with the MC model. The number of sub-classes is reduced by 50\% (or more) in all three dataset balancing cases. The sub-class removal approach leads to a slightly fewer sub-classes than the clustering approach. 

\begin{figure}[]
    \vspace{2mm}
    \centering
\includegraphics[width=0.8\linewidth]{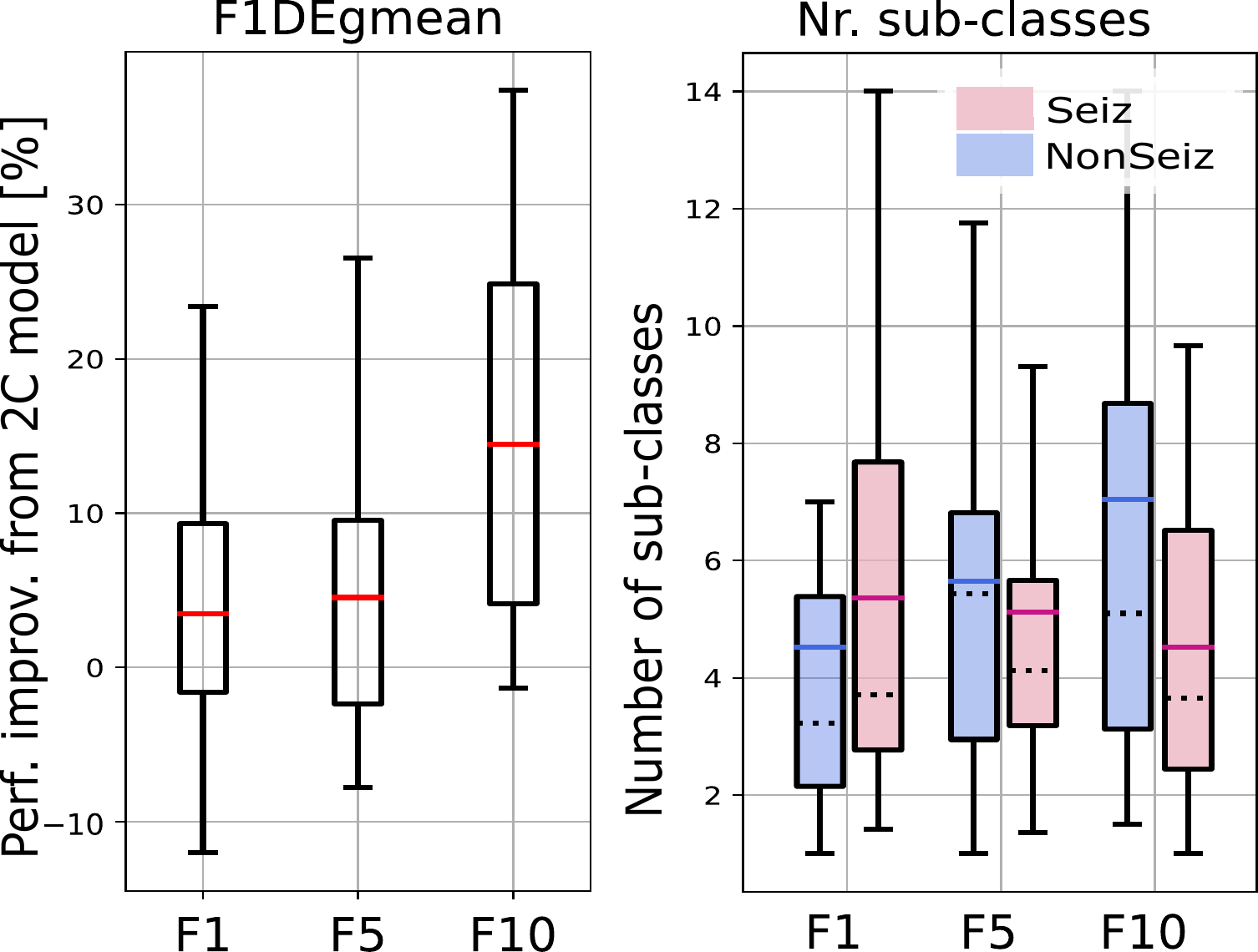}
    \caption{\small{Summarized results for multi-class learning when compared to 2-class learning. Performance improvement and number of sub-classes is reported for all three balancing datasets.}} 
    \label{fig:ResultsSummary_graph2}
\end{figure}

In Fig.~\ref{fig:ResultsSummary_graph2}, we summarize our experimental results and show the final performance improvement and the number of sub-classes after the multi-centroid model with the removal of sub-classes (MCred) for all three balancing datasets. Performance improvement is the smallest for F1, as the space for improvement is also the smallest. For highly unbalanced data (F10), an increase of up to 14\% (on top of initial performance) for the test set was achieved. The multi-centroid approach has the biggest potential for more unbalanced datasets, which are closer to real-life data distribution and also have the lowest absolute performance. As seen in Fig.~\ref{fig:ResultsSummary_graph1}, F1 performance is improved from ~80\% to ~83\% for the test set after smoothing, and from ~60\% to ~73\% for F10. 

When observing the number of sub-classes for different dataset balancing strategies, it seems that the more non-seizure data considered, the more sub-classes are necessary. This conclusion is logical, as we add more data that can represent different neural activity states. On the other hand, in terms of seizure sub-classes, the more non-seizure data we have, the fewer seizure sub-classes we need, as training is less sensitive to small changes in seizure dynamics. Thus, our results indicate that the multi-centroid approach is more significant the closer we are to more realistic data balancing.

\section{Conclusion}

In this work, we have presented a novel semi-supervised learning approach aimed at improving hyperdimensional computing models. The multi-centroid approach was tested on the challenging use case of epileptic seizures detection. In particular, based on given global labels (seizure or non-seizure), instead of forcing only two HD prototype vectors, one for each class, we allow unsupervised creation of any number of sub-classes and their centroid vectors (of seizure and non-seizure). This enables less common signal patterns not to be under-represented, but to create their own sub-class when they are significantly different from the existing sub-classes. 

Our proposed multi-centroid approach has significantly improved performance when compared to a simple 2-class HD model; up to 14\% on the test set of the most challenging dataset with 10 times more non-seizure than seizure data.  
It also leads to the creation of a highly variable number of seizure and non-seizure sub-classes for each subject, reflecting the complexity of the data and the classification challenge itself. One drawback of this approach is the memory requirements that storing all sub-class model vectors implies. However, this increment is linear with the number of sub-classes and can be easily constrained according to the hardware requirements of different types of possible final wearable platforms.

Then, we designed and tested two approaches for optimizing the number of sub-classes, while still keeping an improved performance, as well as sub-classes reduction and clustering. Both approaches have led to a significant reduction of the number of sub-classes ($\sim$50\%), while maintaining equally high performance as the first expanding step of the initial multi-centroid model.

Finally, the multi-centroid has proven to be able to reach bigger improvement for less-balanced datasets. At the same time, the total number of sub-classes is not significantly increased compared to the balanced dataset. Thus, it can be an important step forward to achieve high performance in epilepsy detection with real-life data distributions, where seizures are infrequent, especially during online learning. 

\printbibliography

\end{document}